\documentclass[conference]{IEEEtran}
\usepackage{tikz}
\usetikzlibrary{decorations.pathreplacing,calligraphy,patterns}
\usepackage{epsfig}
\usepackage{subcaption}
\IEEEoverridecommandlockouts
\usepackage[hyphens]{url}
\usepackage{pifont}
\usepackage{colortbl}
\usepackage{cite}
\usepackage{amsmath,amssymb,amsfonts}
\usepackage{algorithmic}
\usepackage{graphicx}

\usepackage{textcomp}
\usepackage{xcolor}
\usepackage[symbol]{footmisc}
\def\BibTeX{{\rm B\kern-.05em{\sc i\kern-.025em b}\kern-.08em
    T\kern-.1667em\lower.7ex\hbox{E}\kern-.125emX}}
\usepackage[hidelinks]{hyperref} 
\hypersetup{
    colorlinks=True,
    allcolors=black,
    }
\usepackage{cleveref}

\newif\ifdraft
\draftfalse

\makeatletter 
\newcommand{\linebreakand}{%
  \end{@IEEEauthorhalign}
  \hfill\mbox{}\par
  \mbox{}\hfill\begin{@IEEEauthorhalign}
}
\makeatother

\newcommand{\Dist}[0]{\text{Dist}}

\Crefformat{figure}{#2Fig.~#1#3}
\Crefmultiformat{figure}{Figs.~#2#1#3}{ and~#2#1#3}{, #2#1#3}{ and~#2#1#3}
\begin{document}

\title{
Does Knowledge About Perceptual Uncertainty Help an Agent in Automated Driving?
}

 \author{\IEEEauthorblockN{1\textsuperscript{st} Natalie Grabowsky}
 \IEEEauthorblockA{\textit{Institute of Mathematics} \\
 \textit{Technical University of Berlin}\\
 Berlin, Germany \\
 grabowsky@math.tu-berlin.de}
 \and
 \IEEEauthorblockN{2\textsuperscript{nd} Annika Mütze}
 \IEEEauthorblockA{\textit{Department of  Mathematics} \\
 \textit{University of Wuppertal}\\
 Wuppertal, Germany \\
 muetze@uni-wuppertal.de}
 \and
 \IEEEauthorblockN{3\textsuperscript{rd} Joshua Wendland}
 \IEEEauthorblockA{\textit{Faculty of Computer Science} \\
 \textit{Ruhr University Bochum}\\
 Bochum, Germany \\
 joshua.wendland@rub.de} \\
 \linebreakand
 \IEEEauthorblockN{4\textsuperscript{th} Nils Jansen}
 \IEEEauthorblockA{\textit{Faculty of Computer Science} \\
 \textit{Ruhr University Bochum}\\
 Bochum, Germany \\
 n.jansen@rub.de} \and
 \IEEEauthorblockN{5\textsuperscript{th} Matthias Rottmann}
 \IEEEauthorblockA{\textit{Department of  Mathematics} \\
 \textit{University of Wuppertal}\\
 Wuppertal, Germany \\
 rottmann@uni-wuppertal.de}}

\maketitle

\begin{abstract}
Agents in real-world scenarios like automated driving deal with uncertainty in their environment, in particular due to perceptual uncertainty. Although, reinforcement learning is dedicated to autonomous decision-making under uncertainty these algorithms are typically not informed about the uncertainty currently contained in their environment.
On the other hand, uncertainty estimation for perception itself is typically directly evaluated in the perception domain, e.g., in terms of false positive detection rates or calibration errors based on camera images. Its use for deciding on goal-oriented actions remains largely unstudied.
In this paper, we investigate how an agent's behavior is influenced by an uncertain perception and how this behavior changes if information about this uncertainty is available. 
Therefore, we consider a proxy task, where the agent is rewarded for driving a route as fast as possible without colliding with other road users. For controlled experiments, we introduce uncertainty in the observation space by perturbing the perception of the given agent while informing the latter. 
Our experiments show that an unreliable observation space modeled by a perturbed perception leads to a defensive driving behavior of the agent. Furthermore, when adding the information about the current uncertainty directly to the observation space, the agent adapts to the specific situation and in general accomplishes its task faster while, at the same time, accounting for risks.
\end{abstract}


\section{Introduction}
\label{sec:introduction}
\begin{figure}[t]
    \centering
    \begin{subfigure}[t]{0.2\columnwidth}
    \centering
        \includegraphics[width=\textwidth]{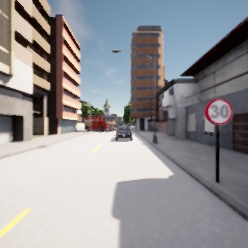}
        \caption{Front\\view}
    \end{subfigure}
    \begin{subfigure}[t]{0.2\columnwidth}
        \centering
        \includegraphics[width=\textwidth,trim= 65 0 65 60,clip]{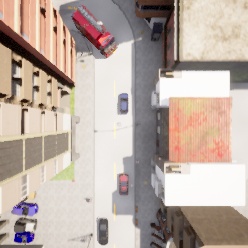}
        \caption{Bird's eye view (BEV)}
    \end{subfigure}
    \begin{subfigure}[t]{0.2\columnwidth}
        \centering
        \includegraphics[width=\textwidth, trim= 65 0 65 60, clip]{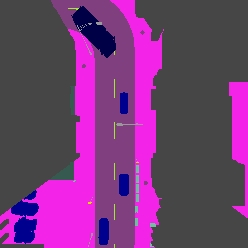}
        \caption{Correct\\semantic segmentation BEV}
    \end{subfigure}
    \begin{subfigure}[t]{0.2\columnwidth}
        \centering
        \includegraphics[width=\textwidth, trim= 65 0 65 60,clip]{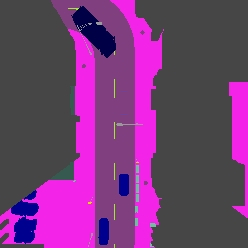}
        \caption{Perturbed semantic segmentation BEV}
    \end{subfigure}
    \caption{An illustration of the main idea of our experiments in the CARLA driving simulator. An agent perceives its environment through a semantic segmentation mask in a bird's eye view, which we perturb in a controlled manner. This corresponds to perfectly quantifiable perceptual uncertainties, which we can provide to the agent. In our study, we investigate whether the agent benefits from this uncertainty information.}
    \label{fig:page1-fig}
\end{figure}

Agents in a real-world scenario perceive their environment by sensors. In the context of autonomous driving (AD), a camera perception is often used as a basis for the learning task performed by an agent.
Typical AD systems can be broadly categorized into two different approaches:
Tasks can be learned end-to-end \cite{bojarski2016endendlearningselfdriving,7995975} 
or the learning process is modularized into consecutive subtasks \cite{learningdrlcarla2021,8917306}. In the former, the raw sensor output is used to predict the next action with the help of, e.g., a reinforcement learning (RL) agent \cite{Perot_2017_CVPR_Workshops,8460934}.  
On the other hand, in modular approaches \cite{8917306}, a perception step, like semantic segmentation \cite{learningdrlcarla2021}, precedes the actual learning.
RL has proven successful for learning tasks based on a reward given a sequence of actions, see \cite{sutton1998rlintro} for an introduction.
In AD, agents can learn to solve challenging tasks like stopping in front of red traffic lights \cite{toromanoff2020end}, lane change \cite{ye2020automatedlanechangestrategy} and trajectory planning \cite{park2024trajectoryplanningautonomousvehicle}.
In general, autonomous vehicles must not only optimize their primary objectives but also ensure compliance with critical safety constraints \cite{sootla2022sauterlsurelysafe}. 
Safety is often addressed by uncertainty-aware RL models \cite{DBLP:journals/corr/KahnVPAL17,8793611}, e.g., by including model uncertainty in the reward \cite{zhang2022safereinforcementlearningcontrastive}, or e.g., by penalizing unsafe trajectories \cite{thomas2021safe}.  
However, most approaches assume accurate observations \cite{zhang2024safetyguaranteedrobustmultiagent}. In contrast, 
AD-systems face sensor issues (e.g., dirty or wet camera lenses) or uncertainty in perception \cite{Holder_2021_ICCV}. Agents that base
their decision-making on this perception may
end up in life-threatening failures. 
Although deep neural networks (DNNs) have shown remarkable performance in computer vision (CV) tasks, visual perception is fraught with uncertainty. Therefore, it is argued that a reliable uncertainty estimation for perception modules is needed \cite{kendall2017uncertaintiesneedbayesiandeep}.
There is a broad variety of methods to estimate the prediction uncertainty \cite{gal2016dropout,lakshminarayananSimpleScalablePredictive2017}
in particular in semantic segmentation \cite{metaseg,maag2024pixelwisegradientuncertaintyconvolutional}. Such approaches are evaluated in terms of false positive detection capabilities \cite{9116288,hendrycks2018baselinedetectingmisclassifiedoutofdistribution}, detection of out-of-distribution (OOD) objects \cite{anthony2023usemahalanobisdistanceoutofdistribution,Chan_2021_ICCV,9564545} or in terms of calibration errors \cite{naeini_mahdi_pakdaman_obtaining_2015}.
In this work, we explore how uncertainty estimates influence RL by examining the learner's behavior \textit{with and without} access to information about the current perceptual uncertainty during driving.
This addresses an open question in RL on how ambiguous and noisy data can be handled by RL approaches \cite{doi:10.1177/0278364913495721}, as well as the open question in CV how relevant uncertainty estimation for perception is in AD.
In other words, our research question is as follows: \emph{Does an RL agent learn to adapt its behavior when informed that the perception is currently uncertain?}

For our experimental analysis, we use the proximal policy optimization (PPO) RL algorithm \cite{ppo} which interacts with a driving simulator environment.
Since RL algorithms optimize their policy by exploring the state space, simulators like CARLA \cite{carla} are used to simulate real-world scenarios.
The proxy learning task in our experiments is deliberately chosen to be simple as our focus does not lie on learning complex tasks but in detecting changes in policy when an agent is informed of uncertain perception. 
The agent's objective is to drive straight from a start to an end point as fast as possible without colliding with other road users by accelerating or reducing its speed.
The agent perceives its environment based on a bird's eye view (BEV) semantic segmentation of the current stretch of road. To simulate uncertainty in the perception, we (temporarily) remove individual road users in the segmentation mask, exemplary shown in \Cref{fig:page1-fig}.
The segmentation mask and the uncertainty information are encoded in the observation space.
Our experiments show that the RL agent informed about the uncertainty of the perception can adapt to the situation to a certain extent.
We summarize our contributions as follows:
\begin{itemize}
    \item We provide an experimental setup for RL under an uncertain segmentation-based perception. This uncertain perception is complemented with indicators of uncertainty, mimicking optimal uncertainty estimation. While our setup is simplistic, it addresses major challenges of automated driving.
    \item We investigate the role of perceptual uncertainty estimation in enhancing the downstream RL task of AD.
    \item Our experiments reveal that unreliable perception typically results in defensive driving behavior. However, when perceptual uncertainty is incorporated into the observation space, the agent learns to adapt more effectively to changing situations.
\end{itemize}
Our code is publicly available at 
\url{https://github.com/nagrab/Does-Knowledge-About-Perceptual-Uncertainty-Help-an-Agent-in-Automated-Driving}

\section{Related Work}
With our study, we investigate the interface between RL and perceptual uncertainty estimation. Therefore, we a) discuss uncertainty estimation in RL and b) examine uncertainty estimation in CV and its applications.

\paragraph{Uncertainty Estimation in RL}
When applying RL models in the real world, safety constraints need to be fulfilled \cite{10675394}. Uncertainty estimation or uncertainty aware-models \cite{DBLP:journals/corr/KahnVPAL17,8793611,zhang2022safereinforcementlearningcontrastive,thomas2021safe} can be used to assess safety guarantees. 
This can help to avoid unsafe trajectories \cite{thomas2021safe}, states 
\cite{pmlr-v37-sui15,wachi2018safeexploration} or actions 
\cite{zhang2022safereinforcementlearningcontrastive}. Gaussian processes are used to model unknown functions like the reward function to restrict the exploration to only safe states \cite{pmlr-v37-sui15,wachi2018safeexploration}. 
Additionally, the estimated uncertainty information can be included into the reward \cite{zhang2022safereinforcementlearningcontrastive} or cost function \cite{DBLP:journals/corr/KahnVPAL17}. 
Zhang and Guo propose a risk preventive training method which allows to choose trajectories with a low risk based on an uncertainty estimate for a state-action pair leading to unsafe states \cite{zhang2022safereinforcementlearningcontrastive}. In the approach proposed by Kahn et al.\ the cost function depends on the estimated collision probability which leads to a defensive behavior in unknown environments \cite{DBLP:journals/corr/KahnVPAL17}.
Furthermore, the uncertainty of decisions can be modeled by Bayesian RL techniques like ensembles networks \cite{hoel2020tactical,Zhang2024SafeReinforcmentLearning}.
These approaches aim at finding a safe policy function or value function by performing uncertainty-aware policy optimization, however most of the approaches assume an accurate observation of the environment.
Nevertheless, perception and raw sensor data may be uncertain or error-prone \cite{kendall2017uncertaintiesneedbayesiandeep}.
In contrast to the aforementioned studies, we investigate how the agent's behavior is influenced when it is informed about the presence or absence of uncertainty in the perception.

\paragraph{Uncertainty Estimation in CV}
For scene understanding based on semantic segmentation, several uncertainty estimation (UE) methods were proposed for perception networks \cite{gal2016dropout,efron1987better}. 
Monte-Carlo dropout \cite{gal2016dropout,kendall2017uncertaintiesneedbayesiandeep}, approximating Bayesian inference, is one of the simplest but effective methods for UE, however inefficient due to multiple inferences per image. 
Shen et al.~\cite{shen2021real} propose a distillation method to learn the conditional predictive distribution of a dropout model, thus establishing real time capability. 
The knowledge about the uncertainty can  be used e.g., to detect OOD objects \cite{maag2024pixelwisegradientuncertaintyconvolutional}, detect false positive predictions \cite{9116288} or calibrating confidences \cite{naeini_mahdi_pakdaman_obtaining_2015}. However, UE methods are often limited to one specific application. 
In the study of Kahl et al.~\cite{kahl2024valuesframeworksystematicvalidation}, the authors point out a gap between theory and practice of UE methods. They propose that uncertainty methods should be evaluated on multiple relevant downstream tasks as varied as OOD-detection \cite{Oberdiek_2020_CVPR_Workshops,Cen_2021_ICCV,ancha2024deep}, active learning \cite{colling2020metaboxnewregionbased,tejaswi2019regionbasedal,pmlr-v235-franco24a} and failure detection \cite{9116288,yingda2020synthesizethencompare,chan2019metafusioncontrolledfalsenegativereduction}.
However, all approaches have in common that the evaluation of the UE methods is limited to vision tasks. Its application to RL is sparsely researched.
To our knowledge, the effect of informing the agent about uncertainty in the perception module 
has yet to be addressed.
With our work we address this interface between vision uncertainty and its influence on subsequent actions of an RL agent.

\section{Construction of the Study}
\label{ch:method}
In this section, we describe the setup to investigate RL under an uncertain segmentation-based perception. We define a state space $S$, representing the true state of the environment, an observation space $O$, which encodes the agent’s perceptual information, and an action space $A$. The relationship between the true state and the agent’s observations is governed by the observation function  $\Omega \colon S \times A \to \Dist(O)$,
which maps to the set of all distributions over the observation space $O$.
A reward function is given by $R \colon S \rightarrow \mathbb{R}$ and a discount factor by $\gamma \in (0,1]$.
A policy is a map $\pi \colon O \rightarrow A$.
We consider a finite horizon specified by discrete time steps $T = \{t_1, \dots, t_\text{max}\}$.
An agent's objective is to find a policy that maximizes the expected discounted cumulative reward, as in $\max_\pi \mathbb{E} \left[ \sum_{t=t_1}^{t_\text{max}} R(s_t) \gamma^t \right]$.
\vspace{2pt}

We continue by introducing the general pipeline and then describe the action, state and observation space in more detail. Finally, we explain how we simulate the uncertainty in the observation space as well as model the reward. 

\subsection{Scenarios of Perception and its Uncertainty (Estimation)}
\label{ch:architecture}
\usetikzlibrary{fit,shapes.arrows}
\newcommand{\cmark}{\ding{51}}%
\newcommand{\xmark}{\ding{55}}%
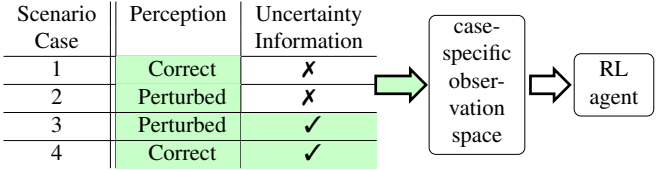
\begin{figure}
\centering
\resizebox{\columnwidth}{!}{%
\begin{tikzpicture}
    \definecolor{hellgruen}{RGB}{200,255,200}

    \node (tabelle) {
        \begin{tabular}{c||c|c}
            Scenario & Perception & Uncertainty \\ 
            Case &  & Information \\ \hline
            1 & \cellcolor{hellgruen} Correct & \xmark \\ \hline
            2 & \cellcolor{hellgruen} Perturbed & \xmark \\ \hline
            3 & \cellcolor{hellgruen} Perturbed & \cellcolor{hellgruen} \cmark \\ \hline
            4 & \cellcolor{hellgruen} Correct & \cellcolor{hellgruen} \cmark \\
        \end{tabular}
    };

    \coordinate (a) at (2.9,0);
    \coordinate (b) at (3.5,0);
    \coordinate (c) at (3.2,0);
    \coordinate (d) at (5.45,0);

    \coordinate (e) at (5.3,0);
    \coordinate (f) at (5.73,0);
    
    \node[single arrow, draw=black, very thick, fill=hellgruen,
      minimum width = 10pt, single arrow head extend=3pt,
      inner xsep=0pt,
      fit=(a) (b)] {};

    \node[single arrow, draw=black, very thick, fill=white,
      minimum width = 10pt, single arrow head extend=3pt,
      inner xsep=0pt,
      fit=(e) (f)] {};

    \node[
        draw=black,
        rounded corners=3pt,
        fill=white,
        align=center,
        anchor=west,
        xshift=0.5cm,
        text width=1.25cm
    ] at (c) {case-specific observation space};

    \node[
        draw=black,
        rounded corners=3pt,
        fill=white,
        align=center,
        anchor=west,
        xshift=0.5cm,
        text width=1.0cm
    ] at (d) {RL agent};

\end{tikzpicture}
}
\caption{We consider four scenarios differing in perturbation of the observation space as well as the availability of the uncertainty information. Green color indicates the scenario-dependent components of the observation space.}
\label{fig:expsetup}
\end{figure}
The generated driving scenarios can be classified by 1) the perception being either correct or perturbed and 2) the agent either being informed about the perturbation or not. This results in four possible combinations that are provided as the input of the observation space, as shown in \Cref{fig:expsetup}.
In the case of a perturbation being present, we provide the exact corresponding uncertainty which we refer to as the \emph{uncertainty information} in the following.

The agent perceives its environment through colorized semantic segmentation masks depicting the stretch of road with the agent vehicle from a bird's eye perspective.
Depending on the presence of uncertainty, the original segmentation image or a perturbation thereof is encoded in the observation space.
For the scenarios where the agent has knowledge (or information) about the perception's uncertainty, the observation space is enlarged to encode this information as well.
The agent can either brake or throttle to avoid colliding with other road users while driving the route as fast as possible.

To model a realistic driving scenario, we use a continuous action and state space. 
For learning, we use the on-policy PPO algorithm \cite{ppo}.
We consider the first task proposed in \cite{carla}, driving a straight road, but enlarge the complexity by including dynamic objects. For our study, we refrain from steering but include a  
braking action to enable an active decision on reducing the speed to not collide with a preceding vehicle. 

\subsection{Action Space}
\label{ch:action}
We define the action space $A = [-1, 1]$, where an action $\Tilde{a}_t \in A$ for a time step $t \in T$ is considered to be accelerating for $\Tilde{a}_t \geq 0$ and braking for $\Tilde{a}_t < 0$.
The accelerating or braking intensity is given by $| \Tilde{a}_t |$, where 0 is no intensity and 1 is full intensity.
However, the environment is not updated with the momentum-free action $\Tilde{a}_t$, but with a modified action $a_t$ which we yield by using an inertia model $\rho$ as in

\begin{equation}
    \begin{aligned}
        a_t & = \rho(\Tilde{a}_t, a_{t-1}) \\
        &= 
        \begin{cases} 
            0.9 \, \Tilde{a}_t & \text{if } \operatorname{sgn}(\Tilde{a}_t) \neq \operatorname{sgn}(a_{t-1}), \\
            0.9 \, \Tilde{a}_t + 0.1 \, a_{t-1} & \text{else.}
        \end{cases}
    \end{aligned}
    \label{eq:action}
\end{equation}
The inertia model simulates more realistic driving dynamics and essentially dampens the chosen action.

\subsection{State Space}
\label{ch:statespace}

\begin{figure}
    \centering

 \resizebox{\columnwidth}{!}{\begin{tikzpicture}
 \draw[fill = teal, opacity=.4] (-6,0.5) rectangle (-4,-0.5) node[pos=0.5,text opacity=1] {\large Sensor};
 \draw[thick] (-4,0.5)--(-2.75,4.0);
 \draw[thick] (-4,-0.5)--(-2.75,-4.0);
        \node at (-2,0) {\includegraphics[width=0.07\textwidth]{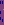}};
        \draw[color=red] (-2,-3) rectangle (-1.7,-2.05);
        \draw[->] (-1,-2.425)--(-1.7,-2.425) node[right, color=red, pos=-0.2] {\large Agent};
        \draw[decorate, decoration = {brace}, rotate=180] (1.35,4)--(2.65,4) node[below,pos=0.5] {\large $4$};
        \draw[thick,decorate,decoration= {brace}, rotate=180] (2.7,4)--(2.7,-4) node[left,pos=0.5] {\large $25$};
        \node at (2,0) {\includegraphics[width=0.07\textwidth]{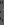}};
        \node at (4,2.75) {\includegraphics[width=0.0065\textwidth, trim = 0 80 0 0, clip]{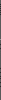}};
        \node at (4,-2.75) {\includegraphics[width=0.0065\textwidth, trim = 0 0 0 80, clip]{Images/Observationspace/flatte_scale.png}};
        \node at (4,0.25) {\vdots};
        \node at (4,1.25) {\vdots};
        \node at (4,-1.2) {\vdots};
        \node at (4,0.75) {\vdots};
        \node at (4,-.25) {\vdots};
        \node at (4, -.75) {\vdots};
        \draw[thick,->] (-1.35,0)--(1.39,0) node[above, pos=0.5] {\large Gray scaled};
        \node[rotate=90] at (3.3,0) {\large Flatten}; 
        \draw[pattern = crosshatch dots gray,fill = gray] (6,3.75)--(6.5,3.75)--(6.5,-2.25)--(6,-2.25)--(6,3.75);
        \draw[pattern color=gray!50,
pattern=north east lines,fill = cyan] (6,-2.25)--(6.5,-2.25)--(6.5,-3.25)--(6,-3.25)--(6,-2.25);
        \draw[fill=blue] (6,-3.75)rectangle(6.5,-3.25);
        \draw [thick,decorate, decoration={brace,amplitude=10pt}] (6.5,-3.25)--(6.5,-3.75);
        \node[text width=4.5cm] at (9.5,-3.375) {\large Uncertainty component\\ ($4$ element)};
        \draw[thick,->] (4.2,0)--(6,0);
        \draw[thick,decorate, decoration= {brace, amplitude=10pt}] (6.5,3.75)--(6.5,-2.25) ;
        \node[ text width=4.5cm] at (9.5,.75) {\large Vision component\\ ($100$ elements)};
        \draw[thick,decorate, decoration= {brace,amplitude=10pt}] (6.5,-2.25)--(6.5,-3.25);
        \node[ text width=4.5cm] at (9.5,-2.375) {\large Non-visual component\\ ($6$ elements)};
    \end{tikzpicture}}
\caption{Observation space setup. The observation space dimension increases when the uncertainty is added. We distinguish between three components: 1) Vision component -- consisting of a flattened and gray-scaled semantic segmentation image; 2) Non-visual component - containing, e.g., velocity; 3) Uncertainty component -- containing the one-hot encoded uncertainty information.}
    \label{fig:statespace}
\end{figure}

The state space $S$ holds all the information of the environment, including position and velocities of all vehicles.
Additionally, it includes the uncertainty information describing whether an observation is perturbed or not.

\subsection{Observation Function and Observation Space}
The observation function $\Omega$ maps the state $S$ directly into the observation space $O$.
A resulting observation $o_t \in O$ of a given state $s_t \in S$ at a given time point $t \in T$ may contain only partial and also perturbed information of $s_t$.
The observation space is defined by three parts as
\[
    O = C_\text{vision} \times C_\text{non-vis} \times C_\text{uncertainty},
\]
and is visualized in \Cref{fig:statespace}.
The components contain distinct semantical information for the agent.
The first component holds the visual information -- in form of a segmentation mask -- originating from the BEV above the agent obtained in CARLA~\cite{carla}.
It is defined as a $100$-dimensional vector $C_\text{vision} = \{0, \dots, 255 \}^{100}$ for the $256$ shades of the gray-scaled mask, and a dimension of $100$, as the segmentation mask's dimensions are $4 \times 25$. 
The second part is defined as $C_\text{non-vis} = \mathbb{R}^6$, which contains the ``non-visual'' information such as the throttle and braking intensity, velocity, normalized velocity, the orientation and distance to the lane center, following the convention of \cite{grundlagenpaper}.
Finally, the remaining part contains the uncertainty information and is defined as $C_\text{uncertainty} = \{0, 1\}^4$.
This 4-dimensional vector stores the one-hot encoded information about which of the four perceptual perturbations the agent is currently experiencing (\Cref{fig:expsetup}).
However, for experiments where no uncertainty information is provided to the agent, we define the component as $C_\text{uncertainty} = \{()\}$, a singleton set therefore holding no information.
A single observation 
is given by $o_t = (v_t, d_t, u_t)$, where $v_t$ is the flattened segmentation mask, $d_t$ the ``non-visual'' information and $u_t$ the uncertainty information.

\subsection{Perturbing the Perception}
\begin{figure}
    \centering
    \begin{tikzpicture}
        \draw[dotted] (-3,-0.25)--(2.5,-0.25) node[pos=.5,above] {\small a: No next front vehicle (VEXV)};
        \draw[rounded corners,color=cyan,fill] (-3,-0.3) rectangle (-2,-0.6) node[pos=0.5,color=black] {$b$};
        \draw[rounded corners, color = pink, fill] (-1.5,-0.3) rectangle (-.5,-.6) node[pos=.5,color=black] {ego};
        \filldraw[ rounded corners,pattern color=green,  draw = black, pattern=north east lines,opacity=.5] (0,-.3) rectangle (1,-.6) node[pos=.5,color=black] {$f_1$};
        \draw[rounded corners, color=violet,fill] (1.5,-.3) rectangle (2.5,-.6) node[pos=.5,color = black] {$f_2$};
        
        \draw (-3,-.65)--(2.5,-.65);
        \draw[dotted] (-3,-1.2)--(2.5,-1.2) node[pos=.5,above] {\small b: No following vehicle (XEVV)};
        \filldraw[ rounded corners,pattern color=cyan,  draw = black, pattern=north east lines,opacity=.5] (-3,-1.25) rectangle (-2,-1.55) node[pos=0.5,color=black] {$b$};
        \draw[rounded corners, color = pink, fill] (-1.5,-1.25) rectangle (-.5,-1.55) node[pos=.5,color=black] {ego};
        \draw[rounded corners, color=green,fill] (0,-1.25) rectangle (1,-1.55) node[pos=.5,color=black] {$f_1$};
        \draw[rounded corners, color=violet,fill] (1.5,-1.25) rectangle (2.5,-1.55) node[pos=.5,color = black] {$f_2$};
        \draw(-3,-1.6)--(2.5,-1.6);

        \draw[dotted] (-3,-2.15)--(2.5,-2.15) node[pos=.5,above] {\small c: No front vehicles (VEXX)};
        \draw[rounded corners, color=cyan,fill] (-3,-2.2) rectangle (-2,-2.5) node[pos=0.5,color=black] {$b$};
        \draw[rounded corners, color = pink, fill] (-1.5,-2.2) rectangle (-.5,-2.5) node[pos=.5,color=black] {ego};
        \filldraw[ rounded corners,pattern color=green,  draw = black, pattern=north east lines,opacity=.5] (0,-2.2) rectangle (1,-2.5) node[pos=.5,color=black] {$f_1$};
         \filldraw[ rounded corners,pattern color=violet,  draw = black, pattern=north east lines,opacity=.5](1.5,-2.2) rectangle (2.5,-2.5) node[pos=.5,color = black] {$f_2$};
        \draw(-3,-2.55)--(2.5,-2.55);

        \draw[dotted] (-3,-3.1)--(2.5,-3.1) node[pos=.5,above] {\small d: No vehicles (XEXX)};
         \filldraw[ rounded corners,pattern color=cyan,  draw = black, pattern=north east lines,opacity=.5]  (-3,-3.15) rectangle (-2,-3.45) node[pos=0.5,color=black] {$b$};
        \draw[rounded corners, color = pink, fill] (-1.5,-3.15) rectangle (-.5,-3.45) node[pos=.5,color=black] {ego};
        \filldraw[ rounded corners,pattern color=green,  draw = black, pattern=north east lines,opacity=.5] (0,-3.15) rectangle (1,-3.45) node[pos=.5,color=black] {$f_1$};
         \filldraw[ rounded corners,pattern color=violet,  draw = black, pattern=north east lines,opacity=.5](1.5,-3.15) rectangle (2.5,-3.45) node[pos=.5,color = black] {$f_2$};
        \draw(-3,-3.5)--(2.5,-3.5);

         \draw[dotted] (-3,-4.05)--(2.5,-4.05) node[pos=.5,above] {\small e: All vehicles (VEVV)};
         \draw[rounded corners, color=cyan, fill] (-3,-4.1) rectangle (-2,-4.4) node[pos=0.5,color=black] {$b$};
        \draw[rounded corners, color = pink, fill] (-1.5,-4.1) rectangle (-.5,-4.4) node[pos=.5,color=black] {ego};
        \draw[rounded corners, color=green,fill] (0,-4.1) rectangle (1,-4.4) node[pos=.5,color=black] {$f_1$};
        \draw[rounded corners, color=violet,fill] (1.5,-4.1) rectangle (2.5,-4.4) node[pos=.5,color = black] {$f_2$};
        \draw(-3,-4.45)--(2.5,-4.45);
    \end{tikzpicture}
      \caption[Perceptual perturbations in experiments]{A summary of the perceptual perturbations used in our experiments. Four different perturbations are used for simulating uncertainty in perception. The last case stands for a correct perception. We use the abbreviations for visible (V), agent/ego-vehicle (E) and invisible (X).}
    \label{fig:cases}
\end{figure}
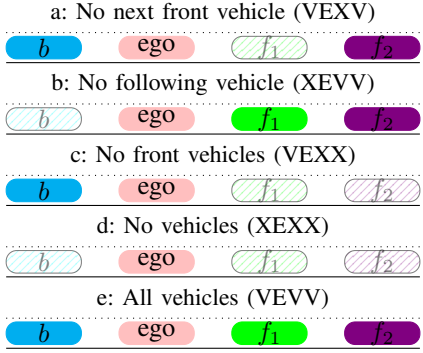
\label{ch:uncertainty_methods}
For the behavioral study, we construct artificial uncertainty cases by perturbing the presence of other vehicles in the visual component $v_t$ of the agent's perception $o_t$.
The agent's vehicle is referred to as ego-vehicle, and we restrict perturbations to the ego-vehicle lane.
Given that at most four vehicles can occur in the stretch of road the agent can perceive, we define five conceivable \textit{perturbations cases} which are listed in
\Cref{fig:cases}. The agent perceives the vehicles in its environment in three different categories: First, the vehicle is visible (V), secondly the vehicle is removed from the mask and therefore not visible (X) or the vehicle is the ego-vehicle (E) itself. For the first case (VEXV) the closest front vehicle ($f_1$) is overlooked. In the second case (XEVV), the following vehicle ($b$) is not visible. The third case (VEXX) occurs when both front vehicles ($f_1$ and $f_2$) are invisible and the fourth case (XEXX) simulates that all but the ego-vehicle are overlooked. In the final case (VEVV) all vehicles are visible. This corresponds to a correct perception.
The one-hot encoding of the uncertainty information for the first case (VEXV) is $u_t = [1,0,0,0]$ and for the unperturbed case (VEVV) is $u_t = [0,0,0,0]$.

\subsection{RL Reward Modeling}
\label{ch:reward}

Most RL experiments in CARLA (see e.g., \cite{grundlagenpaper,8917306,learningdrlcarla2021}) model the reward function as a linear combination of state variables such as velocity and do not consider time.
We include a time component to encourage the agent to fulfill its objective as quickly as possible.
This time decay component
weights the momentary speed in direction of the end point.
Our reward $r_t = R(s_t)$ is defined as
\begin{equation}
    r_t = \begin{cases}
        \left(\beta+\Tilde{\beta}\left(1-\frac{t}{t_{\text{max}}}\right)\right) \, v_{\text{mom}} &\,\text{if } t<t_{\text{max}}\\
        -\alpha &\,\text{on failure}\\
        \Tilde{\alpha} &\,\text{if done}.
    \end{cases}
    \label{eq:betabetatildereward}
\end{equation}

All constants $\beta$, $\Tilde{\beta}$, $\alpha$, $\Tilde{\alpha} \in \mathbb{R}^+$. 
The value $t_{\text{max}}\in T$ defines the maximal episode length and $v_\text{mom} \in \mathbb{R}$ is the agent's momentary velocity in direction of the end point.
A negative reward of $- \alpha$ is returned if the agent fails to accomplish its task -- determined as $t_\text{terminal} \in T$ -- by either
1) causing a collision, 2) exceeding the maximal time limit ($t \geq t_\text{max}$) or 3) not start driving soon enough ($t \geq t_\text{bound} \in T$), with $t_\text{max} \geq t_\text{terminal} \geq t_\text{bound}$. 
A positive reward of $\Tilde{\alpha}$ is returned if the agent finishes its task successfully.

The momentary speed is defined as
\begin{equation}
v_{\text{mom}}=\lVert \text{loc}_{f}-\text{loc}_{t-1}\rVert - \lVert \text{loc}_{f}-\text{loc}_{t}\rVert
\label{eq:vmom}
\end{equation}
whereby $\text{loc}_{t}$ is the current location of the agent, and $\text{loc}_f$ the final/target location, that the agent has to reach to finish its task. Equation \eqref{eq:vmom} can be viewed as the velocity in the direction of the final location, normalized by the traveled distance within a single time step.

While the objective of an RL agent is to maximize the expected discounted cumulative sum over all rewards, we analyze the cumulative reward until a time point $t'$ before the termination of an episode, $t' < t_\text{terminal}$.
Therefore, with the discount factor $\gamma=1$ and together with \eqref{eq:betabetatildereward}, the cumulative reward is given by

\begin{equation}
     \sum_{t=t_1}^{t'} \left(\beta+\Tilde{\beta}\left(1-\frac{t}{t_{\text{max}}}\right)\right)v_{\text{mom}},
    \label{eq:return}
\end{equation}
and can be understood as a discrete convolution. 
Since the time component linearly decays over time, its convolution with the momentary velocity is maximal when the latter is high as early as possible. Whenever the agent waits a few time steps, the momentary velocity of subsequent time steps receives less weight.
However, the agent needs to reduce its speed or wait if the front vehicle drives slowly. The factor $\beta$ rewards the momentary speed regardless of time, to account for situational slower driving.
The reward prevents the agent from waiting until the front vehicles pass the end point $\text{loc}_f$ which could make the perception irrelevant for solving the task.

\subsection{Problem Statement}
We pose the following research question: 
Given an observation  $o_t = (v_t, d_t, u_t)$, where $v_t$ is a visual input affected by perturbations,  $d_t$  contains ``non-visual'' driving state information, and  $u_t$  is a one-hot encoded uncertainty information vector specifying the type of perturbation, does providing  $u_t$ 
change the agent's behavior 
compared to an agent that does not receive uncertainty information in terms of failure rate, traveled distance and used time steps within an episode as well as the distance to the front vehicle and brake frequency?

\section{Numerical Experiments}
\label{ch:experiments}
In our behavioral study, we analyze and compare RL agents trained on scenarios 1, 2 and 3 (cf.\ \Cref{fig:cases}). Our key experiment is the one using scenario 3, while the other ones serve for comparison. Note that scenario 4 serves only for testing, but not for learning since the uncertainty is non-informative in that case. For reference, we start showing that the agent finishes its task frequently in the case of a correct perception. The second experiment analyzes the behavior of the agent with perturbed perception but without knowledge about the perception's uncertainty, i.e., scenario no.\ 2. In the last experiment, we analyze the agent's behavior when the agent's perception is perturbed, i.e., uncertain, and the observation space is extended by the uncertainty information, i.e., an agent trained in scenario 3.

\subsection{Description of the Learning Task}
\begin{figure}[t]
\centering
\resizebox{0.98\columnwidth}{!}{\begin{tikzpicture}
        \node at (0,0) {\includegraphics[trim= {0 17cm 90 0}, clip,width=\columnwidth]{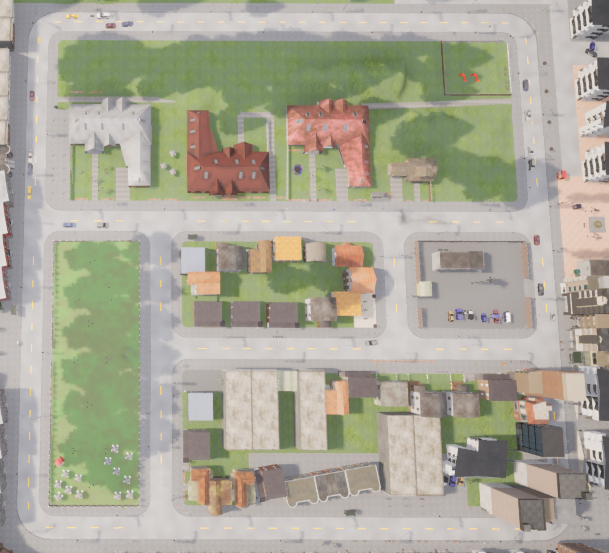}};
      \draw[red, thick] (-3.,0)--(3.5,0);
       \draw[orange, fill] (-3,0) circle (3pt) node[below, black] {Start};
        \draw[orange, fill] (3.5,0) circle (3pt) node[below, black] {End};
    \end{tikzpicture}}
       \caption{Top down view of the agent's task route.}
    \label{fig:taskmap}
\end{figure}
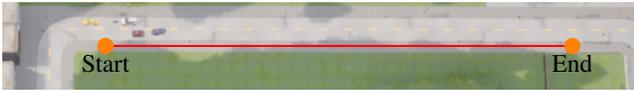

The behavioral analysis is conducted using the open-source simulator CARLA~\cite{carla}, version 0.9.15. The agent's task is to drive $150$ meters straight in Town 2 as presented in \Cref{fig:taskmap}.
In each episode, vehicles are spawned at predefined locations on the map including two vehicles in front of the ego-vehicle.
The front vehicles are intentionally slowed down by periodic braking to enforce adapted (re)actions of the agent.
For each time step, measured in CARLA time steps, the agent obtains the reward defined in \eqref{eq:betabetatildereward} with a scaling factor for the momentary speed of $\beta=3$ and a scaling factor for the time decay of $\Tilde{\beta}=2$.
The task is episodic, i.e., as soon as a failure or finish event is triggered, an episode terminates, the agent is reset and a new episode and therefore a new trail to solve the task is started.
The objective is that the agent finishes its task in at most $t_{\text{max}}=7500$ time steps without any collision. 
Otherwise, the episode terminates and the agent obtains a penalty of $r_t = -\alpha = -50$. 
Additionally, the agent must drive at least $3$ meters in $t_{\text{bound}}=500$ time steps, otherwise the episode terminates directly and the agent receives a penalty of $r_t = -\alpha=-50$. 
If the agent finishes its task successfully, i.e., reaching the end of the route within the specified time period, it obtains a (sparse) reward of $r_t = \Tilde{\alpha}=100$. 

\subsection{Implementation Details}
 
We build upon the on-policy PPO algorithm~\cite{ppo} implementation by Razak and László \cite{grundlagenpaper} and adapted the
action space, the state space, the observation space and the reward according to our described needs. Based on multiple experiments, we adjusted the hyperparameters of the underlying PPO-RL, such that the discounting factor is set to $\gamma=0.999$ and a clip value of $\varepsilon=0.2$. 
Furthermore, we use a learning rate of $10^{-5}$ and for the loss of the PPO a value loss scale of $0.5$ and an entropy scale of $0.01$. 
Our exploration factor is initialized with a value of $\sigma_{0}=0.1$ and each $5\cdot10^{5}$ time steps it is reduced linearly to $\sigma = (\sigma-0.025)$.
Even though PPO provides a probabilistic policy, we made it deterministic by choosing the most likely action in each time step.
\subsection{Overview of Training, Validation and Testing}
For each of the experiments, we first train the agent for two million time steps on its training scenario, select the best model based on validation and then test the agent's behavior under the perceptual perturbations. 
During validation/testing the exploration factor $\sigma$ is set to zero, such that the actions are sampled according to the expectation value of the policy. During the training, each hundredth episode, we validate the agent's policy for $20$ episodes. Additionally, after the training we validate for $20$ episodes, the three episodes where the agent achieved the highest cumulative sum over all rewards
during training. Among those, we chose the policy for our behavioral analysis where the agent achieved the highest cumulative sum over all rewards on average during validation.
We test for $60$ episodes the safety critical cases a), c), d) and e) and a mixture of all cases, referred to as mixed perturbation case (MPC) in the following.
More specifically, the cases a) -- e), uniformly chosen at random with replacement, occur for a time interval of a length randomly sampled from $\{50,100,150,200,400\}$.

\begin{figure}
    \centering
    \begin{subfigure}[t]{0.465\columnwidth}
    \centering
    \includegraphics[width=\textwidth, trim=15 15 15 0, clip]{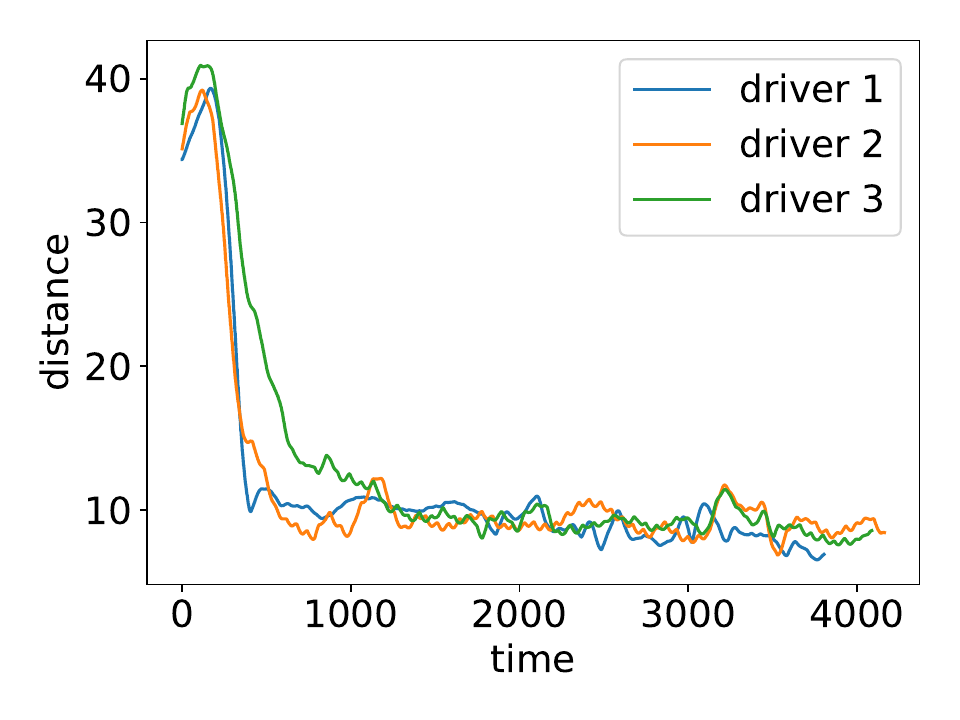}
    \caption{Human driving behavior for scenario 1 illustrated by the distance to the front vehicle}.
    \label{fig:human}
    \end{subfigure} \hfill
    \begin{subfigure}[t]{0.465\columnwidth}
    \centering
        \includegraphics[width=\textwidth, trim=25 0 0 0, clip]{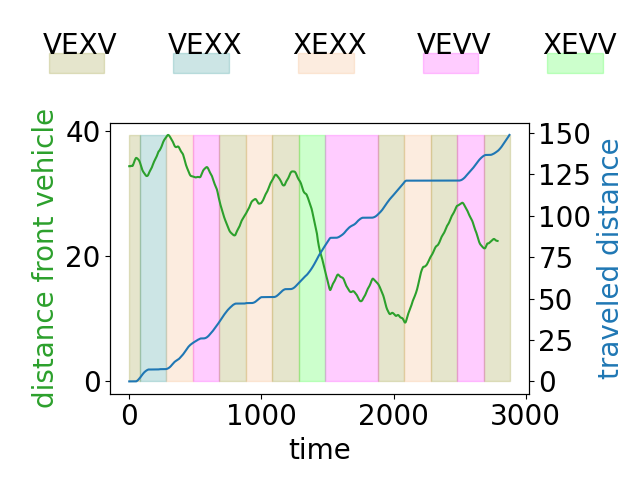}
        \caption{Agent's driving behavior during one random sampled episode in MPC case trained in experiment 3.}
        \label{fig:example}
    \end{subfigure}
    \caption{Left: Human driving behavior in our simulator settings. Right: An example of the agent's driving behavior during the MPC case with fixed time steps for switching the case.}
\end{figure}

The agent's behavior is analyzed based on four different key observations: the traveled distance within one episode, braking to throttle ratio, number of time steps per episode and distance to the front vehicle. \Cref{fig:human} shows an example of the driving behavior of three human drivers when driving with a correct perception, i.e., scenario 1. 
They need some seconds to react and accelerate the vehicle. Thereafter, the distance to the front vehicle ranges between $5$ to $12$ meters mimicking the oscillating behavior of the front vehicle.

\subsection{Experiment 1 -- \textbf{Correct} Perception \textbf{Without} Uncertainty Information}
\label{ch:nounc}

\begin{figure}
\begin{minipage}{0.495\columnwidth}
 \centering
    \includegraphics[width=\textwidth]{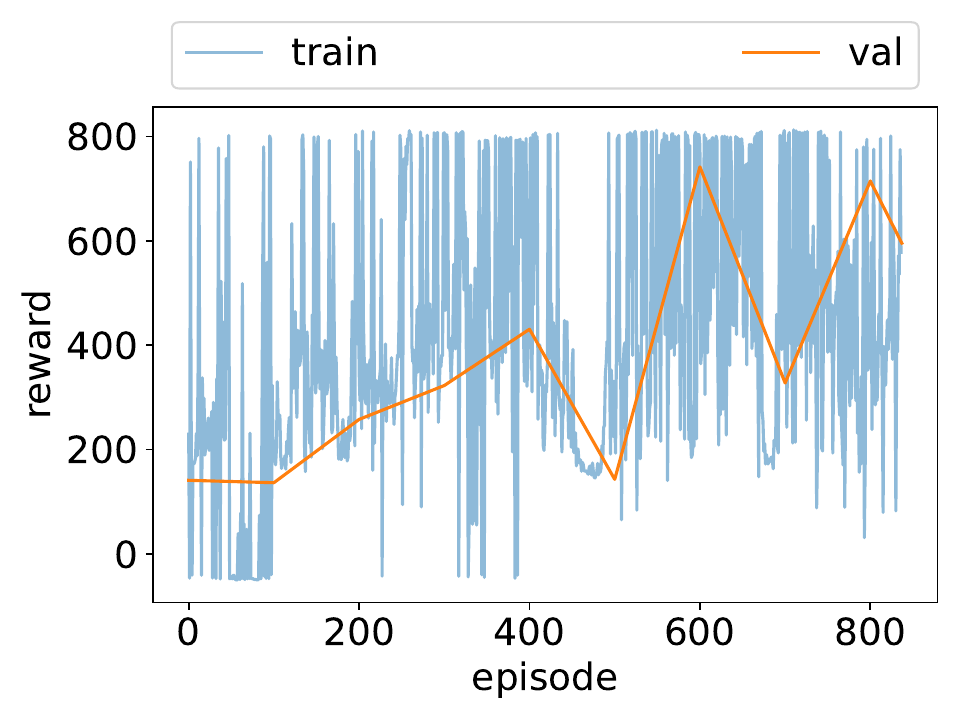}
\end{minipage}
\begin{minipage}{0.495\columnwidth}
   \centering
    \includegraphics[width=\textwidth]{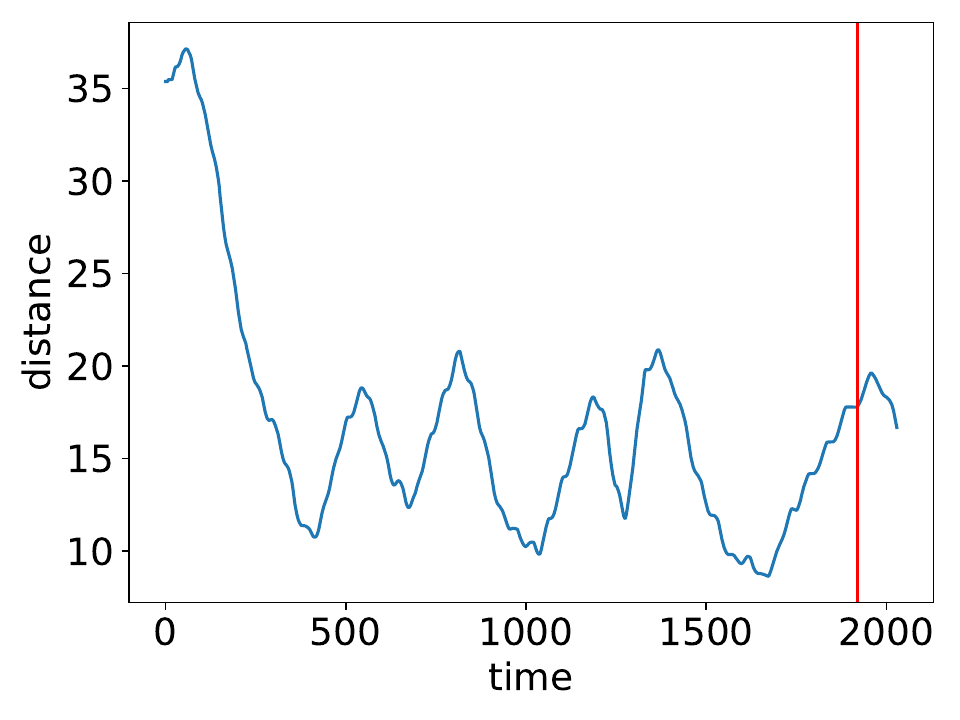}
\end{minipage}
\caption{Left: Reward progression of the training and validation for scenario 1 - correct perception without uncertainty information.
Right: Distance to the front vehicle during one episode. }
  \label{fig:reward_unc}
   \label{fig:distance_nounc}
\end{figure}

Firstly, we examine the agent's behavior when learning under correct perception. Hence, we are in scenario 1 and train an agent on the unperturbed case e) in \Cref{fig:cases}. The observation space does not include any uncertainty information as it is uninformative. 
\Cref{fig:reward_unc} shows the reward progress during training and validation. After about $500$ episodes, the agent finishes its task frequently.
\Cref{fig:distance_nounc} shows a randomly chosen episode tested in the unperturbed case (VEVV) for illustrating the agent's behavior w.r.t.\ the distance to the front vehicle.
Since the agent needs to take action to start driving, but the front vehicle starts driving directly, we observe the small increase in the distance right at the beginning. Later, the agent catches up and the distance decreases. It
follows the front vehicle within 
$20$ to $10$ meters distance. 
\Cref{fig:boxplots_comparison} presents the behavior averaged across all tested episodes for all scenarios and experiments. For experiment 1, the agent finishes its task frequently when the perception is not perturbed.

If at least one of the front vehicles is invisible (VEXV, VEXX), the agent always collides with the front vehicle. This is not surprising, but indicates that the agent bases its behavior on the vision component of the observation space. 
For the perturbed cases where at least one front vehicle is invisible (VEXV,VEXX,XEXX), the agent performs `throttle' most of the time. Also the velocity is higher than for the unperturbed case. Thus, the agent cannot handle a perturbed perception.
\subsection{Experiment 2 -- \textbf{Perturbed} Perception \textbf{Without} Uncertainty Information}
\label{ch:uncnounc}
\begin{figure}[t]
\begin{minipage}{0.495\columnwidth}
        \centering
    \includegraphics[width=\textwidth]{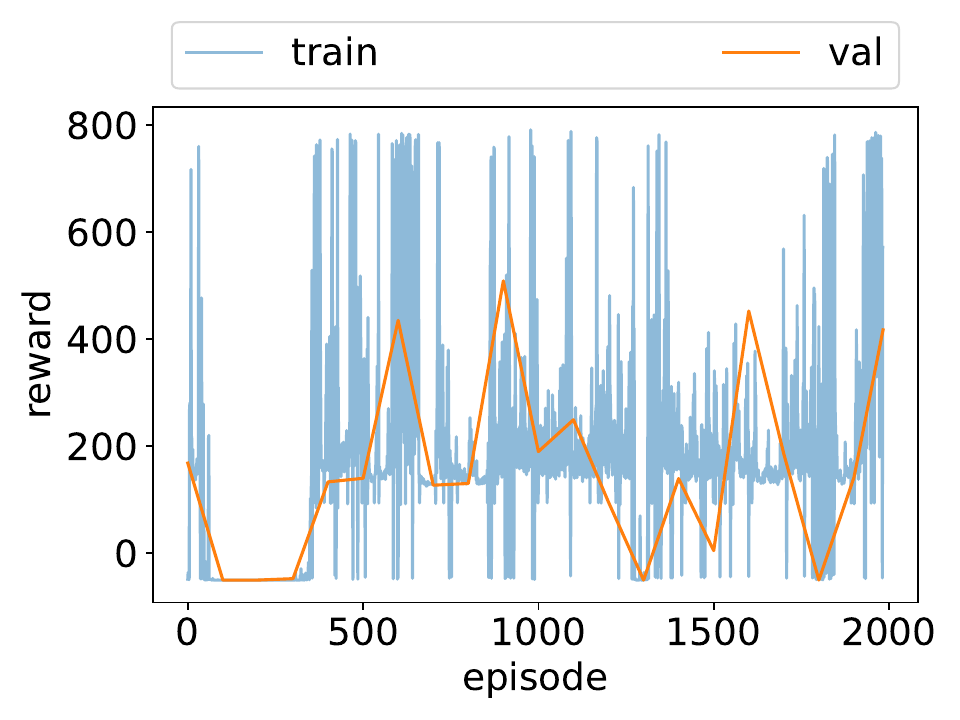}    
\end{minipage}
\begin{minipage}{0.495\columnwidth}
       \centering
    \includegraphics[width=\textwidth]{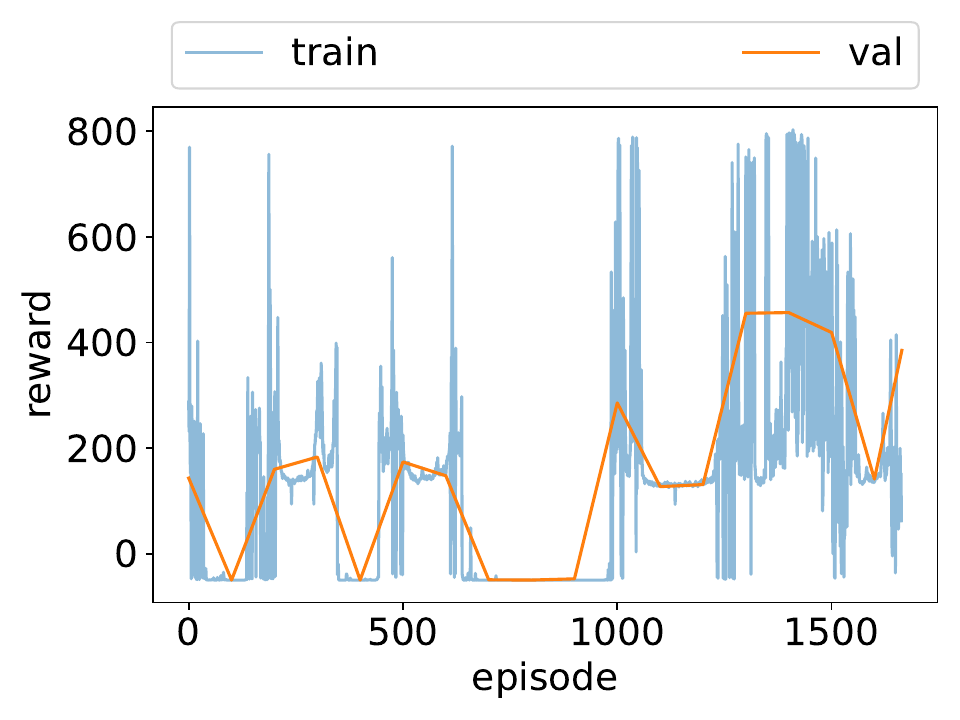}
    
\end{minipage}
\caption{Left: Reward progression when training in scenario 2 with visual perturbation but no uncertainty information. Right: Reward progression when training in scenario 3 with uncertainty information.}
\label{fig:reward_uncnounc}
\label{fig:reward_uncunc}
\end{figure}

In the second experiment, the agent's perception is perturbed but the uncertainty information is not part of the observation space, i.e., we train in scenario 2 using the MPC case. 
The procedure of training, validation and testing is the same as in \Cref{ch:nounc}. \Cref{fig:reward_uncnounc} shows the reward progression during the training and validation. In contrast to experiment 1, the agent occasionally finishes its task during training, despite the uncertainties/perturbations in perception.

\begin{figure*}[t]
    \centering
    \begin{subfigure}{1.11\textwidth}
    \centering
        \begin{minipage}{.7\textwidth}
        \raggedright
        \includegraphics[width=1.15\textwidth, trim=0 10 0 0,clip]{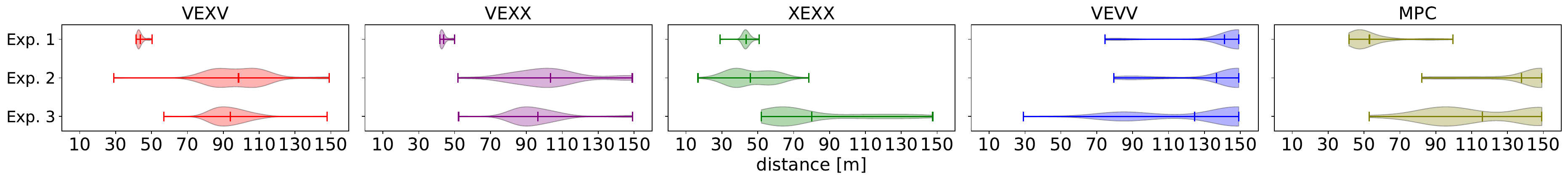}
        \end{minipage}
        \hspace*{-.3cm}
        \begin{minipage}{.3\textwidth}
        \centering
        \caption{Traveled\\ distance}    
        \end{minipage}
    \end{subfigure}
    
    \begin{subfigure}{1.11\textwidth}
    \centering
    \begin{minipage}{.7\textwidth}
    \raggedright
        \includegraphics[width=1.15\textwidth, trim=0 10 0 0, clip]{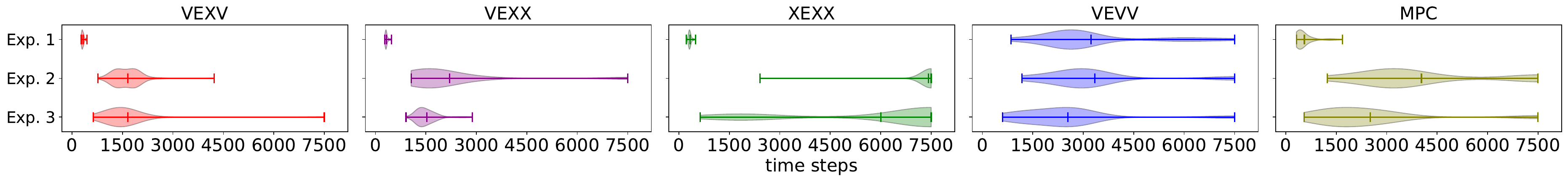}      
    \end{minipage}
    \hspace*{-.3cm}
    \begin{minipage}{.3\textwidth}
    \centering
        \caption{Time \\steps}
    \end{minipage}
    \end{subfigure}
    
    \begin{subfigure}{1.11\textwidth}
    \centering
    \begin{minipage}{.7\textwidth}
    \raggedright
        \includegraphics[width=1.15\textwidth, trim = 0 10 0 0,clip]{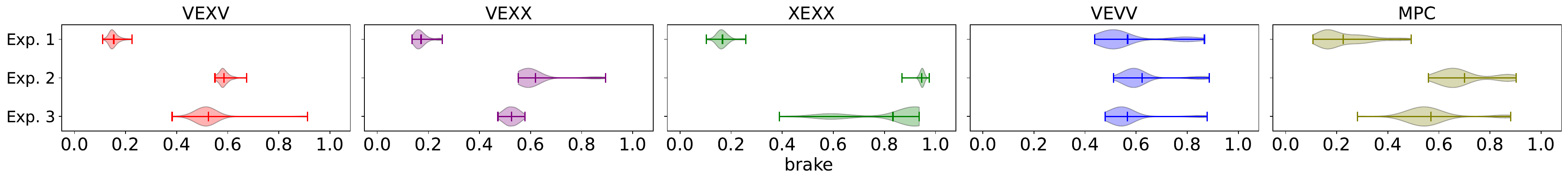}
    \end{minipage}
    \hspace*{-.3cm}
    \begin{minipage}{.3\textwidth}
    \centering
           \caption{Brake\\ fequency} 
    \end{minipage}
       
    \end{subfigure}
    
    \begin{subfigure}{1.11\textwidth}
    \centering
    \begin{minipage}{.7\textwidth}
        \raggedright
        \includegraphics[width=1.15\textwidth, trim= 0 10 0 0, clip]{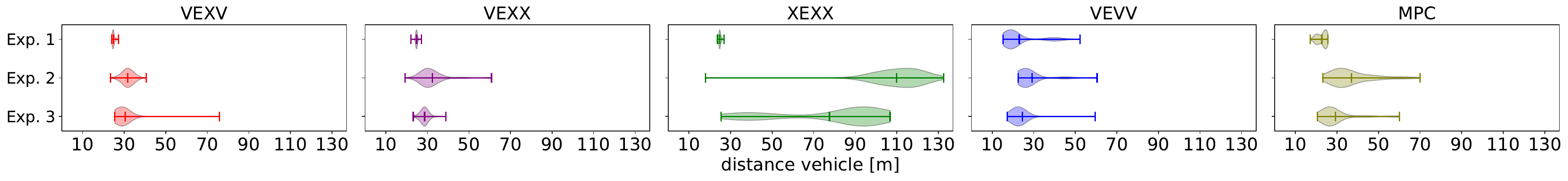}
    \end{minipage}
    \hspace*{-.3cm}
    \begin{minipage}{.3\textwidth}
\centering
        \caption{Distance\\ to front  \\vehicle}
    \end{minipage}
    \end{subfigure}
    \caption{The agent's behavior in terms of distance traveled, time steps per episode elapsed, frequency of brake action and, the distance to the front vehicle. These are presented for the perturbation case (indicated in the titles) and the experiments 1--3.}
    \label{fig:boxplots_comparison}
\end{figure*}

The agent's behavior for the different cases is presented in \Cref{fig:boxplots_comparison}.
The agent finishes the task frequently (in about $70\%$ of its attempts) for the unperturbed case (VEVV). 
In comparison to experiment 1, the agent learns a defensive behavior.
However, for the cases where at least one front vehicles is invisible (VEXX,VEXV,XEXX) the agent either collides with its front vehicles or waits until $t_\text{max}$ is reached.
In all perturbed cases, the agent performs the action `brake' more frequently than in the unperturbed case. Additionally, when comparing distances to the closest front vehicles, the defensive agent (experiment 2) keeps more distance to the front vehicle on average. The velocity is lower for all cases. 

\subsection{Experiment 3 -- \textbf{Perturbed} Perception \textbf{With} Uncertainty Information}
\label{ch:uncunc}
In our final and main experiment, the agent's perception is perturbed and the uncertainty information is added to the observation space, i.e., we are in scenario 3 and train on the MPC case.
The previous two experiments serve for comparison.
\Cref{fig:reward_uncunc} (right-hand panel) presents the reward progression of the training and validation. 
For the MPC case (the training case), the agent finishes its task in $33.33\%$ of the test trials. 

\Cref{fig:boxplots_comparison} summarizes the agent's behavior for all experiments and all tested cases. We discuss the subfigures (a) to (d) consecutively. The traveled distance (subfigure a) in experiment 3 does not show a clear tendency compared to experiment 2, however it is mostly superior to experiment one, except for the unperturbed VEVV case, which is to be expected. In the MPC test case, we see that the agent of experiment three travels slightly less far than the agent of experiment 2, which can be seen as a slightly less defensive behavior in comparison to the agent of experiment 2. The traveled distance (subfigure b), although the distributions differ from subfigure (a), essentially shows the same tendency when comparing all three experiments. Considering the brake frequency (subfigure c) further supports the thesis that the defensiveness has decreased from experiment 2 to 3. Consistently across all test cases, the agent brakes less in experiment 3 than in experiment 2. The distance to the front vehicle (subfigure d) reveals that the agent in experiment 3 tends to keep less distance to the front vehicle than the agent from experiment 2, in particular in the MPC case where the agent of experiment 3 has been trained on. This shows that the agent of experiment 3 has improved capabilities to adapt to the current perturbation in the perception.

We now take a deeper look into the behavior of an agent trained in experiment 3. \Cref{fig:example} shows the behavior of the agent during one episode of the MPC case where the perturbation case varies during the episode. It can be seen again that the agent adapts its behavior to the different cases. However, we note a certain inertia in terms of reaction time needed to adapt to the current perturbation case.
The agent starts driving slowly across the perturbed cases. After $500$ time steps the agent receives the information that its perception is now correct and decreases the distance to the front vehicle. This is followed by some perturbed cases, where at least one front vehicle is invisible. The agent adapts to those cases and increases the distance slowly. Between $1500$ and $1800$ time steps, the agent knows, that only the perception w.r.t.\ the following vehicle is perturbed (XEVV), but not w.r.t.\ the front vehicle. 
In this case, the agent reduces its distance to the front vehicle to about $10$ and $15$ meters. In general, these observations demonstrate the adaptivity of the agent to spontaneous changes of perception quality.
We provide in the Appendix numerical results for all three experiments. 
\section{Conclusion}
\label{ch:conclusion}
In this paper, we studied how the knowledge about perceptual uncertainty, encoded as an uncertainty information in the observation space of an RL agent, influences the learned actions of the agent. 
The task was in the scope of AD, driving $150$ meters straight as quickly as possible without colliding with other vehicles. 
We trained the agent in three different scenarios: 1) correct perception, 2) perturbed perception without uncertainty information, and the main experiment 3) perturbed perception with uncertainty information. Indeed, we found that the agent is able to finish its task in all three scenarios when tested on the training scenario. When evaluated on a perturbed perception but trained with a correct one, the agent fails consistently. The agent trained on perturbed perception without uncertainty information acts in general defensive to avoid collisions. The agent trained on perturbed perception with uncertainty information adapts to the given scenario. Under a perturbed perception it acts more defensively while increasing the pace when the uncertainty information states that the perception is correct. 

\section*{Acknowledgments}
This work is supported by the German Fed.\ Ministry of Education and Research through project ``UnrEAL'', grant no.\ 01IS22069, by the German Fed.\ Ministry of Economy and Climate Action via ``NXT GEN AI METHODS'', grant no.\ 19A23014Q and by the ERC Starting Grant DEUCE, grant no.\ 101077178. We thank the human drivers for their reference drives.
\bibliographystyle{ieeetr}
{\bibliography{main}}
\textbf{\appendix}
The following section discusses the numerical results of the main paper. We discuss the experimental results of experiment 1 (\cref{tab:behaviornouncnounc}, experiment 2 (\cref{tab:behuncnounc}) and experiment 3 (\cref{tab:behuncunc}) and their comparison between experiment 1 and 2 (\cref{tab:compuncnouncnouncnounc}) and between experiment 2 and 3 (\cref{tab:compuncuncuncnounc}). The comparison between experiment 2 and 3 is disregard, since there are two different components: different training style and different observation space, which will have an influence in the comparison. 
\section*{Behavior of the Agent With Different Observation Spaces}
\begin{table*}[tbh]
    \centering
    \caption[Behavior of the agent for VEVV case training without uncertainty information]{Quantitatively values for the behavior analysis of Experiment 1 --  Correct perception without uncertainty information. Trav. dist. shows the traveled distance in meters. Brake describes the frequency how often the action was brake. Time describes the average length of an validation episode in time steps. The distance to the next front vehicle is illustrated in dist.\ front veh.. The last column describes the average velocity per time step.}
    \resizebox{1.75\columnwidth}{!}{\begin{tabular}{|c|c|c|c|c|c|}\hline
        Case & Trav.\ dist.\ [m] & Brake [\%] & Time [steps] & Dist.\ front veh.\ [m]& Vel.\ [tr.\ di./time]   \\ \hline
        VEXV & $\;43.80\pm\;2.03$ & $15\pm\;\,\,2$ & $\;\;327\pm\;\;\;\;43$ & $24.90\pm0.67$ & $0.1339$ \\
        VEXX & $\;43.76\pm\;1.95$ & $17\pm\;\,\,3$ & $\;\;329\pm\;\;\;\;44$ & $24.82\pm0.61$ & $0.1330$\\
        XEXX & $\;43.42\pm\;3.36$ & $17\pm\;\,\,3$ & $\;\;329\pm\;\;\;\;46$ & $24.90\pm0.61$& $0.1320$\\
        VEVV & $141.28\pm20.75$ & $57\pm12$ & $3227\pm1667$ & $22.99\pm8.75$ & $0.0438$\\
        MPC & $\;\;53.00\pm12.51$ & $22\pm\;\,\,9$ & $\;\;552\pm\;\;289$ & $22.70\pm2.27$ & $0.0960$\\\hline
    \end{tabular}}
    
    \label{tab:behaviornouncnounc}
\end{table*}
\Cref{tab:behaviornouncnounc} and \Cref{tab:behuncnounc} show the numerical results of the behavior of the agent trained on different observation spaces without the information about uncertainty as defined in \Cref{ch:statespace}. In addition to the traveled distance, brake ratio, number of time steps and distance to the front vehicle we present the average velocity in the tables. For the first experiment trained with correct perception without accessing the uncertainty information (cf. Section 4.4 main paper) the agent finishes its task in $85\%$ of all validation episodes. Comparing the safety critical test cases VEXV, VEXX, XEXX (cf. \Cref{fig:cases}) and MPC case to the safe case (VEVV, trained scenario) the respective behavior significantly changes. 
\begin{table*}[tbh]
    \centering
    \caption[Behavior of the agent for perturbed and non perturbed state space training]{Quantitatively values for the behavior analysis of Experiment 2 -- Perturbed perception without uncertainty information. Trav. dist. shows the traveled distance in meters. Brake describes the frequency how often action was brake. Time describes the average length of an validation episode in time steps. The distance to the next front vehicle is illustrated in dist.\ front veh.. The last column describes the average velocity per time step.}
   \resizebox{1.75\columnwidth}{!} {\begin{tabular}{|c|c|c|c|c|c|c|}\hline
    Case & Trav.\ dist.\ [m] & Brake [\%] & Time [steps] & Dist.\ front veh.\ [m] & Vel [tr.\ di./time]\\\hline
    VEXV&  $\;\;98.49\pm 18.20$ & $58\pm\;\;2$ & $1654\pm\;\;484$ & $\;\;31.71\pm\;\;3.14$ & $0.0596$\\
     VEXX & $103.46\pm21.88$ & $62\pm\;\;8$ & $2202\pm1673$ & $\;\;32.36\pm\;\;7.01$ & $0.0470$ \\
     XEXX & $\;\;45.90\pm12.28$ & $95\pm\;\;1$ & $7415\pm\;\;652$&$109.94\pm15.23$ & $0.0062$\\
     VEVV& $136.80\pm22.59$ & $62\pm\;\;9$ & $3343\pm1737$ & $\;\;29.15\pm\;\;8.37$ & $0.0409$\\
     MPC & $132.75\pm22.72$ & $70\pm10$ & $3818\pm2009$ & $\;\;38.39\pm11.66$ & $0.0348$\\\hline
    \end{tabular}}
    
    \label{tab:behuncnounc}
\end{table*}
The traveled distance drops at least $55.02$ meters compared to the safe case. Additionally, the brake is used less frequently which increases the average velocity. In total, it can be summarized that the agent drives significantly less defensively than for the safe case and therefore collides more often with the front vehicle. 

Experiment 2 presented in \Cref{tab:behuncnounc} shows the numerical results for the experiment where the agent is trained on the MPC case, i.e.\ perturbed perception, without uncertainty information  and tested under the safety critical cases as well as the safe case. The numerical driving behavior is not significantly different across the cases except for the XEXX case. However, the agent learns a more defensive behavior in contrast to the first experiment which is quantitatively shown in \Cref{tab:compuncnouncnouncnounc}.
\begin{table*}[tbh]
    \centering
    \caption{Comparison between Experiment 2 and Experiment 1. The arrows show in direction of perturbed perception in comparison to correct perception. If $\nearrow$ the value of perturbed perception is larger than the mean plus the standard deviation of the correct one, else if $\searrow$ the values are smaller and if $\rightarrow$ the results are in the same range at least in the standard deviation range. The values are defined as the ratio between the behavior of the agent trained on perturbed perception and the agent trained on the correct perception. The value for Done is the difference between the perturbed perception and correct perception, measured in percentage points.  }
   \resizebox{2\columnwidth}{!}{ \begin{tabular}{|c|c|c||c|c|c|c|c|}\hline
    Case & Reward & Done & Trav.\ dist.\ & Brake & Time & Dist.\ front.\ veh. & Vel \\\hline
    VEXV  &$2.5212$ ($\nearrow$) &$\,\;\;\;\,\,\,3.33$ ($\nearrow$) &$2.2486$ ($\nearrow$) &$3.8667$ ($\nearrow$) &$5.0581$ ($\nearrow$) &$1.2735$ ($\nearrow$) &$0.4451$ ($\searrow$)\\
    VEXX &$2.7199$ ($\nearrow$)&$\;\,\,\,\,10.00$ ($\nearrow$) &$2.3643$ ($\nearrow$) &$3.6471$ ($\nearrow$) &$6.6930$ ($\nearrow$) &$1.3038$ ($\nearrow$) &$0.3534$ ($\searrow$)\\
    XEXX &$0.9750$ ($\rightarrow$) & $\,\,\,\,\;\;\;1.67$ ($\nearrow$) &$1.0571$ ($\rightarrow$) &$5.5882$ ($\nearrow$) &$1.0359$ ($ \nearrow$)& $4.4153$ ($\nearrow$) &$0.0470$ ($\searrow$)\\
    VEVV&$0.9312$ ($\rightarrow$) &$-15.00$ ($\searrow$) &$0.9683$ ($\rightarrow$) &$1.0877$ ($\rightarrow$) &$1.0359$ ($\rightarrow$) & $1.2679$ ($\rightarrow$)&$0.9338$ ($\searrow$)\\
    MPC &$3.0799$ ($\nearrow$) &$\,\,\;\,\,58.33$ ($\nearrow$) &$2.5047$ ($\nearrow$) &$3.1818$ ($\nearrow$) &$6.0167$ ($\nearrow$) &$1.6912$ ($\nearrow$) &$0.3625$ ($\searrow$)\\\hline
    \end{tabular}}
    
    \label{tab:compuncnouncnouncnounc}
\end{table*}
  The comparison is done by using the ratio between the behavior metrics of the agent trained on its perturbed perception (i.e.\ experiment 2) and the behavior metrics of the agent trained on its correct perception (i.e.\ experiment 1) both without information about the perceptual uncertainty. 
  Additionally to the metrics in \Cref{tab:behaviornouncnounc,tab:behuncnounc,tab:behuncunc}, we show the reward and the value, how often each agent finishes its task during validation, called Done. 
  The comparison is done by using the ratio between the behavior metrics of the agent trained on its perturbed perception without the information about the uncertainty (i.e.\ experiment 2) and the behavior metrics of the agent trained on its correct perception (i.e.\ experiment 1).   
  The brake ratio is, except for the safe case, at least three to five times larger than for the experiment where the agent policy is learned based on correct perception. However, for the safe case (VEVV) the measured behavior of the two agents does not show significant differences.
\begin{table*}[tbh]
    \centering
    \caption[Behavior of the agent for different validation classes.]{Quantitatively values for the behavior analysis of Experiment 3 -- Perturbed perception with uncertainty information. The agent is trained on the observation space including perturbed perception and uncertainty information. Trav.\ dist.\ shows the traveled distance in meters. Brake describes the frequency how often the action is brake. Time describes the average length of a validation episode in time steps. The distance to the next front vehicle is illustrated in dist.\ front veh.. The last column describes the average velocity per time step. }
    \resizebox{1.75\columnwidth}{!}{\begin{tabular}{|c|c|c|c|c|c|}\hline
  Case & Trav.\ dist.\ [m] & Brake [\%] & Time [steps] & Dist.\ front veh.\ [m] & Vel.\ [tr.di/time] \\\hline
  VEXV& $\;\;93.88\pm11.66$ & $53\pm\;\;8$ & $1656\pm1112$ & $30.47\pm\;\;8.13$ & $0.0567$\\ 
  VEXX& $\;\;96.33\pm16.32$ & $52\pm\;\;3$ & $1532\pm\;\;353$ & $28.66\pm\;\;2.48$ & $0.0629$\\
  XEXX& $\;\;80.12\pm27.62$ & $83\pm15$ & $6003\pm2465$ & $77.57\pm25.70$ & $0.0133$\\
  VEVV&$124.62\pm32.35$ & $57\pm\;\;9$ & $2546\pm1610$ & $24.43\pm\;\;7.81$ & $0.0489$\\
  MPC&$115.93\pm28.64$ & $57\pm11$ & $2516\pm1799$ & $29.35\pm\;\;8.32$& $0.0489$\\\hline
    \end{tabular}}
    
    \label{tab:behuncunc}
\end{table*}

\Cref{tab:behuncunc} shows the behavior of the agent trained under the MPC case with information about its perceptual uncertainty. For the metrics brake, time and distance to the front vehicle, we observe quantitatively similar results, except for the XEXX case, which is a sign for a case adaptive behavior. For the XEXX case the agent acts highly defensive measured by the long time needed to finish the task, using the brake significantly more often and its overall slow velocity. 

In the following, we compare the results of the agents trained on the perturbed perception where one has information about its perceptual uncertainty whereas the other has not. As before, the comparison is done by using the ratio between the behavior metrics. Here we calculate the ratio of the agent trained on its perturbed perception with the information about the uncertainty (i.e.\ experiment 3) and the behavior metrics of the agent trained on its perturbed perception without the information about the uncertainty (i.e\ experiment 2). The Done value in \Cref{tab:compuncuncuncnounc} is measured by computing the difference between the Done value of experiment 3 and the Done value of experiment 2.
The behavior differs especially for the metric of the average velocity, as well as the distance to the front vehicle. This is shown in \Cref{tab:compuncuncuncnounc}. The training case (MPC) and the safe case (VEVV) show similar results, which could be interpreted as a policy that learns to handle the information and shows that giving the information of safety is not informative. Furthermore, the velocity increases if the uncertainty information is given except for the VEXV case. Also, all other values show the tendency that suggests that the agent drives more adaptively to the situation if it gets access to the uncertainty information.
\begin{table*}[tbh]
    \centering
   \caption[Comparison between validation of training with perturbed state space and with information and no uncertainty information.]{Comparison between Experiment 3 and Experiment 2. The arrows point in direction of the perturbed perception with uncertainty information in comparison to perturbed perception without uncertainty information. I.e., if $\nearrow$ the value of perturbed perception with perceptual uncertainty information is significantly larger than without, else if $\searrow$ the values are smaller and if $\rightarrow$ the results are in the same range at least in the standard deviation range. The values are the ratio between the perturbed perception concatenated with uncertainty information and the perturbed perception without uncertainty information in the observation space. The column Done is the difference between these two validations, measured in percentage point. }
    \resizebox{2\columnwidth}{!}{\begin{tabular}{|c|c|c||c|c|c|c|c|}\hline
        Case & Reward & Done & Trav.\ dist.\ & Brake & Time & Dist.\ front veh.\ &Vel.\  \\\hline
         VEXV&$0.9455$ ($\rightarrow$) &$\;\;-3.33$ ($\searrow$) & $0.9532$ ($\rightarrow$) &$0.9138$ ($\rightarrow$) &$1.0012$ ($\rightarrow$) &$0.9609$ ($\rightarrow$) &$0.9513$ ($\searrow$)\\
         VEXX &$0.9060$ ($\rightarrow$) &$\;\;-8.33$ ($\searrow$) &$0.9311$ ($\rightarrow$) &$0.8387$ ($\nearrow$) &$0.6957$ ($\rightarrow$) &$0.8857$ ($\nearrow$) &$1.3383$ ($\nearrow$)\\
         XEXX &$2.0267$ ($\nearrow$) &$\;\;-1.67$ ($\searrow$) &$1.7451$ ($\nearrow$) &$0.8737$ ($\rightarrow$) &$0.8096$ ($\rightarrow$) &$0.7056$ ($\nearrow$) &$2.1252$ ($\nearrow$)\\
         VEVV&$0.9045$ ($\rightarrow$) &$-10.00$ ($\searrow$) &$0.9110$ ($\rightarrow$) &$0.9194$ ($\rightarrow$ )&$0.7616$ ($\rightarrow$) &$0.8381$ ($\rightarrow$) &$1.1956$ ($\nearrow$)\\
         MPC &$0.8387$ ($\rightarrow$) &$-25.00$ ($\searrow$) &$0.8733$ ($\rightarrow$) &$0.8143$ ($\rightarrow$) &$0.6590$ ($\rightarrow$) &$0.7645$ ($\rightarrow$) &$1.1164$ ($\nearrow$)\\\hline
    \end{tabular}}
    
    \label{tab:compuncuncuncnounc}
\end{table*}
\end{document}